\documentclass{article}

\usepackage[final]{lmca2020}

\usepackage[utf8]{inputenc} 
\usepackage[T1]{fontenc}    
\usepackage{hyperref}       
\usepackage{url}            
\usepackage{booktabs}       
\usepackage{amsfonts}       
\usepackage{nicefrac}       
\usepackage{microtype}      
\usepackage{amsmath}
\usepackage{amssymb,bbm}
\usepackage{graphicx}

\usepackage{multirow}

\newtheorem{theorem}{Theorem}
\newtheorem{lemma}[theorem]{Lemma}

\title{{Learning for Integer-Constrained Optimization through Neural Networks with Limited Training}}
\author{Zhou Zhou, Shashank Jere, Lizhong Zheng, Lingjia Liu}

\author{%
  Zhou Zhou
    \\    
  Bradley Department of ECE, Virginia Tech\\
  Blacksburg, VA, 24060\\
  \texttt{zhouzhou@vt.edu} \\
  \And
   Shashank Jere \\
   Bradley Department of ECE, Virginia Tech \\
   Blacksburg, VA, 24060 \\
   \texttt{shashankjere@vt.edu} \\
   \And
   Lizhong Zheng \\
   Department of EECS, Massachusetts Institute of Technology \\
   Cambridge, MA, 02139 \\
   \texttt{lizhong@mit.edu} \\
   \And
   Lingjia Liu \\
   Bradley Department of ECE, Virginia Tech \\
   Blacksburg, VA, 24060 \\
   \texttt{ljliu@vt.edu} \\
}

\begin{document}

\maketitle
 
\begin{abstract}
In this paper, we investigate a neural network-based learning approach towards solving an integer-constrained programming problem using very limited training. To be specific, we introduce a symmetric and decomposed neural network structure, which is fully interpretable in terms of the functionality of its constituent components. By taking advantage of the underlying pattern of the integer constraint, as well as of the affine nature of the objective function, the introduced neural network offers superior generalization performance with limited training, as compared to other generic neural network structures that do not exploit the inherent structure of the integer constraint. In addition, we show that the introduced decomposed approach can be further extended to semi-decomposed frameworks. The introduced learning approach is evaluated via the classification/symbol detection task in the context of wireless communication systems where available training sets are usually limited. Evaluation results demonstrate that the introduced learning strategy is able to effectively perform the classification/symbol detection task in a wide variety of wireless channel environments specified by the 3GPP community.
\end{abstract}

\section{Introduction}
\label{section:introduction}
Integer-constrained optimization has wide applications in many fields, including industrial production planning, vehicle scheduling in transportation networks, and resource allocation for cellular networks where the variables often represent finite decisions and countable quantities~\cite{glover1986future, hemmecke2010nonlinear}. 
In this paper, we consider developing a neural network (NN)-based solver for the integer-constrained programming problem in the following form:
\begin{equation}
\label{Opt}
\begin{aligned}
& \underset{\boldsymbol x}{\text{minimize}}
& & 
f\left({\boldsymbol H}{\boldsymbol x}-{\boldsymbol y}\right)\\
& \text{subject to}
& & x_{n} \in  {\mathcal A}
\end{aligned},
\end{equation}
where $\boldsymbol H$ is an $Q \times N$ matrix; $\boldsymbol y$ is the vector of dimension $Q \times 1$ containing observed values; $f(\cdot)$ is the objective function; $\boldsymbol x$ is the unknown vector of dimension $N \times 1$ ($x_n$ is the $n$th element of $\boldsymbol x$); and $\mathcal A$ is the set containing the limited integer values that each $x_n$ can take.
Without loss of generality (WLOG), $\mathcal A$ is set to be $\{ -2M-1, -2M+1, \cdots, -1,1,\cdots, 2M-1, 2M+1\}$. This integer-constrained definition of set $\mathcal{A}$ can be generalized to any arbitrary integers. 
Unlike unfolding any optimization based algorithms to a deep neural network~\cite{hershey2014deep}, our approach is motivated by the intrinsic geometry of the integer-constraint ${\mathcal A}$, and the embedded affine-mapping $\boldsymbol H$ in the objective.

Integer-constrained optimization problems are generally NP-hard ~\cite{hemmecke2010nonlinear} due to the non-convex nature of the set ${\mathcal A}^{N}$, where $N$ is the dimension of $\boldsymbol x$. 
An exhaustive search can be prohibitively expensive, especially when $N$ is large. 
Therefore, rather than focusing on conventional optimization-based approaches, we alternatively consider a data-driven or learning-based classification/detection framework where the structural information that is inherent in the constraint can be utilized to improve the learning ``efficiency'' in terms of the training data overhead and ``effectiveness'' in terms of the generalization performance. 

The process of our introduced neural network based solver is as follows:
\begin{itemize}
\item We first define a probability residual-model for ${\boldsymbol r} = {\boldsymbol H}{\boldsymbol x}-{\boldsymbol y}$ where ${\boldsymbol r}$ is generated according to a distribution $P({\boldsymbol r})$ which is determined based on the objective function $f(\cdot)$\footnote{For instance, the Gaussian distribution can be used as $P({\boldsymbol r})$ for L$2$-norm minimization.}. 
\item Then, we generate $K$ tuples of $({\boldsymbol x}, {\boldsymbol r})$ according to the distributions $U({\mathcal A}^N)$ and $P({\boldsymbol r})$, where $U({\mathcal A}^N)$ represents the uniform distribution defined on ${\mathcal A}^N$. The resulting tuples can be organized as,
\begin{align*}
\left\{\left({\boldsymbol x}^{(1)}, {\boldsymbol r}^{(1)}\right), \left({\boldsymbol x}^{(2)}, {\boldsymbol r}^{(2)}\right), \cdots, \left({\boldsymbol x}^{(K)}, {\boldsymbol r}^{(K)})\right)\right\}.
\end{align*}
For a fixed $\boldsymbol H$, $\boldsymbol y$ is generated using $\boldsymbol x$ and $\boldsymbol r$ via the residual relation ${\boldsymbol y} = {\boldsymbol H}{\boldsymbol x} - {\boldsymbol r}$. For the purpose of training a neural network, we have $K$ tuples of $({\boldsymbol x}, {\boldsymbol y})$:
\begin{align}
\label{training_set}
\left\{\left({\boldsymbol x}^{(1)}, {\boldsymbol y}^{(1)}\right), \left({\boldsymbol x}^{(2)}, {\boldsymbol y}^{(2)}\right), \cdots, \left({\boldsymbol x}^{(K)}, {\boldsymbol y}^{(K)})\right)\right\}.
\end{align}
\item We train a neural network $\mathcal N$ using the tuples defined in (\ref{training_set}), where ${\boldsymbol y}^{(k)}$ is the input to $\mathcal N$ and ${\boldsymbol x}^{(k)}$ is the associated output.
\item After $\mathcal N$ is trained, we feed the observation vector $\boldsymbol y$ to $\mathcal N$ to generate a solution, ${\boldsymbol {\hat x}}$, of (\ref{Opt}).
\end{itemize}

To provide a meaningful and practical solution for a large dimensional $\boldsymbol x$, we develop a \emph{parallel} solver in which $N$ independent neural networks simultaneously estimate each entry of $\boldsymbol x$. This is accomplished by leveraging the concept of marginal estimation, tailored specifically to our chosen probability model for the residue $\boldsymbol r$. This decomposed approach can also be extended to obtain a semi-decomposed framework, thus providing a middle ground between this decomposed approach and the joint approach. 

The innovation in the introduced decomposed estimator relies on the geometric structure of the lattice spanned by the affine mapping $\boldsymbol H$ as well as the step-size selected from $\mathcal A$. 
By incorporating such structural information into our neural network design, we can achieve superior generalization performance with very limited training samples. 
The resulting structure of the designed neural network also allows an analytical characterization of the classification/detection error in terms of an asymptotic bound. 
For evaluation, we utilize this optimization framework to perform the symbol detection task in wireless communications systems, which we believe is a meaningful step towards the interdisciplinary research in applying neural networks to solve practical problems of fifth generation (5G) networks~\cite{alliance20155g}.

In summary, the key contributions of this paper are summarized as follows: 
\begin{itemize}
\item The introduced neural network architecture is symmetric and its structure is fully explainable. The neural network aims to learn the marginal probability distribution function (PDF) of each variable. 
Instead of performing an integral to compute the marginal distribution, the introduced method simplifies the posterior estimation by using structured classifiers.

\item With a limited amount of training, the introduced learning approach can achieve better generalization performance than that of generic neural networks having an arbitrary structure, since the structural knowledge of the objective term and of the integer constraints are well integrated in the design of the neural network.

\item We derive a theoretical upper-bound for the generalization error of our introduced neural network. This provides performance guarantees to the introduced neural network  in the classification/detection task.
\end{itemize}

\section{Related Work and Background}
As a special case of the objective function $f(\cdot)$, the quadratic loss/error term has been widely adopted for characterizing residual ``loss'' or ``error'' terms and often appears as the second order term for objective approximation~\cite{mairal2010online, lee2012proximal, parikh2014proximal}. 
When the decision variables are constrained to be integers, the optimization becomes challenging since the integer constraint renders the problem non-convex. 
A wide range of approaches have been introduced for this class of problems, such as branch and bound, convex relaxation, etc. \cite{schrijver1998theory}. 
These approaches can be briefly summarized into the two categories of joint and decomposed methods. 
The joint solver approach operates by directly optimizing the objective function over the parameter grid spanned by the Cartesian product of all variable entries. 
At a given iteration, the variables are adjusted by changing an approximated bound of the overall loss function.
Under this approach, sphere decoding is considered as the near optimal solver in the context of symbol detection for communication systems~\cite{hassibi2005sphere}. 
Theoretically, the joint solver yields the optimal solution after all the branches are checked out. 
However, it places an extremely heavy computational burden, making its scaling to a larger system unfeasible.
The decomposed approach is as an efficient way to address the scalability issue~\cite{caroe1999dual}. 
It essentially avoids the overwhelming complexity of the joint approach by breaking down the aggregated loss term into smaller composite loss terms. 
However, in order to implement the decomposed approach, either an explicit separable form of the objective function or an implicit coordinate transformation is required~\cite{boyd2007notes, price1995pairwise}. 

In this paper, we use a neural network structure that can be deconstructed in order to solve the stated optimization problem.
Our method circumvents the obstacle of the features being non-separable in the objective function and attempts to solve the optimization problem via a pure data-driven framework. 
Related approaches that utilize the decomposed objective function can be found in~\cite{bertsimas2018data,bertsimas2006robust}. 
The introduced learning framework leverages the structural information of the optimal residual probability distribution as introduced in (\ref{training_set}). 
Prior work on using neural networks as probability distribution estimators can be found in~\cite{richard1991neural, price1995pairwise, uria2013rnade, doya2007bayesian, de2003bayesian}. 
However, in these frameworks, inference of the posterior distribution typically requires training a substantially large number of neural network parameters, thereby significantly increasing the training overhead. 
Furthermore, the neural network architectures considered in such frameworks usually do not consider the specific underlying structural information inherent in the problem.

Related work on connecting neural networks to optimization techniques can be found in \cite{ cochocki1993neural, bouzerdoum1993neural,amos2017optnet}. 
In these cases, neural networks are often used to assist the optimization step or reversely, the optimization is considered as part of the neural network operations. 
As an example of the latter, \cite{amos2017optnet} introduced a formulation of integrating quadratic programming as neural network layers. 
The output of the previous layers of the neural network are considered as encoded parameters of the objective function and the associated constraints. Genetic algorithms are a class of random-based evolutionary algorithms that can be applied to constrained optimization problems~\cite{evolutionary2015}. MATLAB's Optimization Toolbox provides an implementation of the genetic algorithm that can be used to evaluate its performance in an integer-constrained optimization task such as the one in Equation~\eqref{Opt}.

The introduced method does not follow the concepts of unrolling or integration as in the existing literature. 
Instead, its motivation is derived purely from a data-driven perspective.
The task chosen to evaluate the introduced neural network architecture is that of classification/transmit symbol detection in wireless communications systems. Symbol detection is the most important receiver processing function in the physical layer (PHY) of a wireless communications system. 
Detecting symbols in vastly different radio environments is one of the major challenges in fifth generation (5G) wireless systems and is critical in guaranteeing the reliability of communication under heterogeneous scenarios. 
In fact, for linear transmission channels with Gaussian noise, the symbol detection task can be exactly formulated as~\eqref{Opt}, where the function $f$ is the L$2$-norm. 
Since the interference plus noise in 5G applications can be highly non-Gaussian, it is desirable to identify solvers that are robust and adaptive to any interference and noise environments.
In neural network-based approaches, a well-trained model can mitigate the non-Gaussian effect of interference from other users that is superimposed on the target signal.
This provides a strong motivation to investigate neural network-based approaches to solve the classification problem which is essentially tied to the integer-constrained optimization task studied in this paper.
\emph{DetNet}~\cite{samuel2019learning} is a state-of-the-art neural network architecture for symbol detection. 
Even though its performance is comparable to conventional model-based methods, it requires a prohibitively large amount of offline training.
\emph{MMNet}~\cite{Khani2020} is another recently introduced neural network-based symbol detection framework for (massive) MIMO (Multiple Input Multiple Output) systems, that utilizes correlation in channel for accelerated training and has a lower computational complexity compared to \emph{DetNet}.
Compared with existing neural network-based methods, we show that our method provides a guaranteed generalization performance with limited training.

\section{Structure-Aware Learning} 
\label{framework}
Based on the probability model presented in (\ref{training_set}), we first consider the joint posterior probability distribution. 
According to Bayes's rule, the joint posterior probability distribution can be written as,
\begin{align*}
& p\left(x_1, x_2, \cdots, x_N|{\boldsymbol y}\right)  \\
& = p\left(x_1 | {\boldsymbol y}\right) p\left(x_2 | x_1, {\boldsymbol y}\right) \cdots p\left(x_N | x_1, \cdots, x_{N-1}, {\boldsymbol y}\right)
\end{align*}
To facilitate a decomposed approach for the maximum a posteriori estimation (MAP), we utilize the following naive Bayesian approximation:
\begin{align}
& \arg \max_{x_1, x_2, \cdots, x_N} p(x_1,x_2,\cdots, x_N|{\boldsymbol y}) \nonumber \\
& \approx \arg \max_{x_1,x_2,\cdots,x_N}p(x_1|{\boldsymbol y})p(x_2|{\boldsymbol y})\cdots p(x_N|{\boldsymbol y}),
\end{align}
indicating that the joint MAP estimator can be approximated as the product of the individual marginal MAP estimators.

\subsection{Binary Integer Constraint}
\label{binary_integer_con}
We now consider how to obtain the marginal MAP estimator $\arg \max_{x_n} p(x_n|{\boldsymbol y})$ using a neural network. For ease of discussion, we first assume that the integer constraint is only restricted to a binary set, i.e., ${\mathcal A} = \{-1, 1\}$. We then learn a neural network ${\mathcal N}_n$ to approximate $p(x_n|{\boldsymbol y})$, where the input of the neural network is $\boldsymbol y$ and the output is the probability mass of $x_n$. Alternatively, we can write the output as the following ratio,
\begin{align}
\label{likelihood_func}
L_{+-}\left({\boldsymbol y}\right): = {\frac{{p\left(x_{n} = 1| \boldsymbol{y}\right)}}{{p\left(x_{n}= -1 | \boldsymbol{y}\right)}}}.
\end{align}
This is termed as the likelihood ratio (LR) between hypotheses $\{H_0: x_n = 1\}$ and $\{H_1: x_n = -1\}$.
Within the neural network, we choose soft-max as the activation function for the final output layer,
\begin{align*}
p\left(x_n = 1 | \mathbf{q}\right) &= \frac{e^{\mathbf{q}^{\top} \mathbf{w}_{1}}}{e^{\mathbf{q}^{\top} \mathbf{w}_{1}} + e^{\mathbf{q}^{\top} \mathbf{w}_{-1}}},\\
p\left(x_n = -1 | \mathbf{q}\right) &= \frac{e^{\mathbf{q}^{\top} \mathbf{w}_{-1}}}{e^{\mathbf{q}^{\top} \mathbf{w}_{1}} + e^{\mathbf{q}^{\top} \mathbf{w}_{-1}}},
\end{align*}
where $\mathbf q$ represents the last hidden state and ${\mathbf w}_1$, ${\mathbf w}_{-1}$ are the weights connecting the hidden layer to the output nodes for $+1$ and $-1$ respectively. We choose cross-entropy as the loss function used in the training of $\mathcal{N}_n$. The training set for ${\mathcal N}_n$ consists of the $K$ tuples:
\begin{align}
\left\{\left({ x}_n^{(1)}, {\boldsymbol y}^{(1)}\right), \left({x}_n^{(2)}, {\boldsymbol y}^{(2)}\right), \cdots, \left({ x}_n^{(K)}, {\boldsymbol y}^{(K)}\right)\right\},
\end{align}
where ${x}_n^{(k)}$ is the $n$th entry of ${\boldsymbol x}^{(k)}$, as defined in (\ref{training_set}). 
\begin{figure}
    \centering
    \includegraphics[height = 0.45\linewidth]{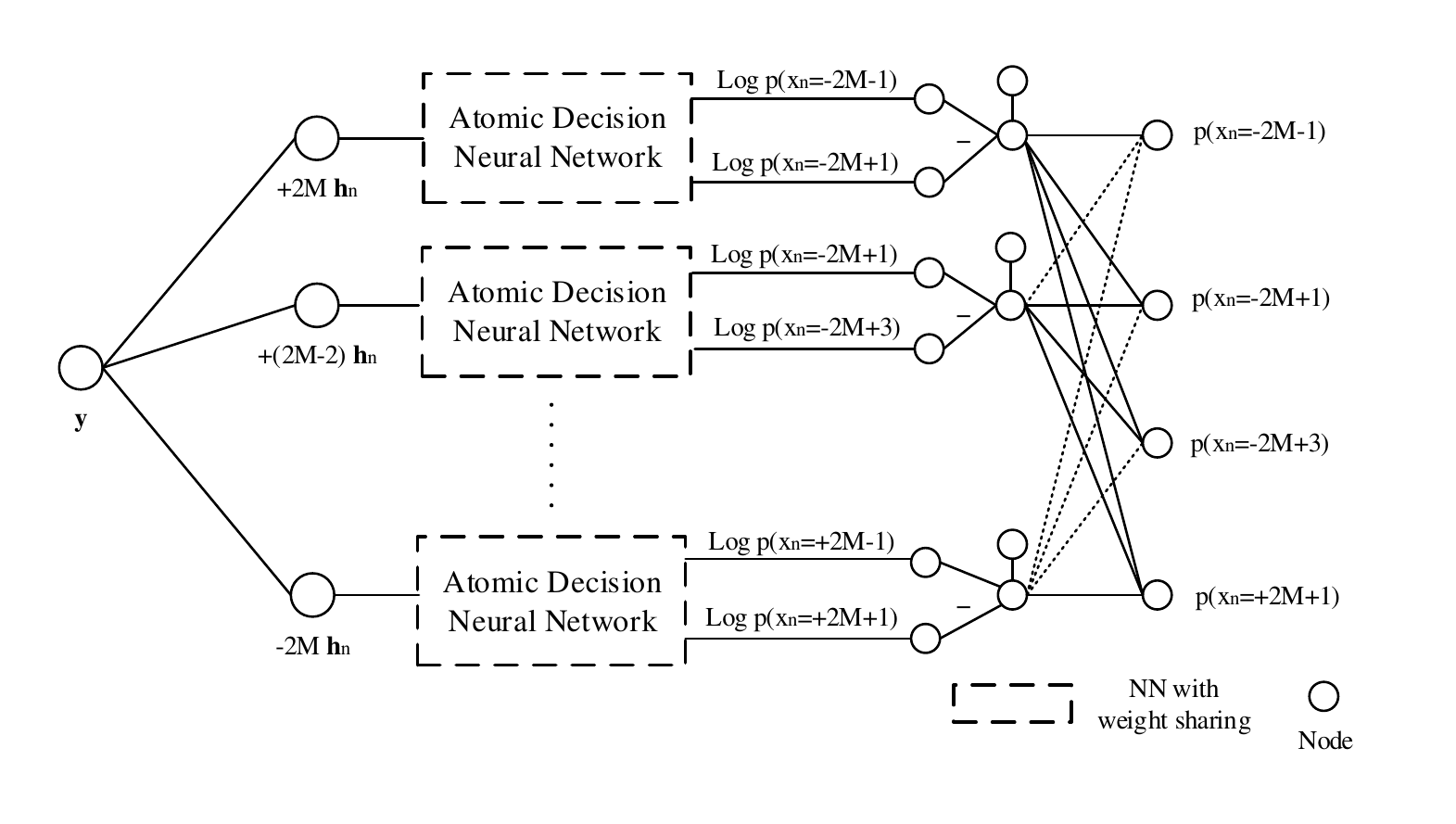}
    \caption{The neural network structure for the posterior estimation of general integer constraint.}
    \label{NN_structure}
\end{figure}

\label{Integer_reg}
\subsection{General Integer Constraint}

Next, we consider the extension of the binary constraint to a general integer constraint, i.e., ${\mathcal A} = \{ -2M-1, -2M+1, \cdots, -1,1,\cdots, 2M-1, 2M+1\}$. Leveraging the pattern of this integer constraint, the neural network ${\mathcal N}_n$ is designed as shown in Fig. \ref{NN_structure}, where the ``atomic decision neural network'' (ADNN) is the same as the binary integer classifier/ MAP estimator introduced in Section \ref{binary_integer_con}. We see that the input of each ADNN is a ``shifted'' version of the observed signal vector $\boldsymbol y$, that is shifted in the direction of ${\boldsymbol h}_n$ (,i.e., the $n$-th column of $\boldsymbol H$), with the size of the shift depending on the step size between the elements of the integer set $\mathcal A$. In other words, for the $m^{\text{th}}$ ADNN, the output ratio represents an estimate of the LR between the hypotheses $\{H_0^{(-2m)}: x_n = -2m+1\}$ and $\{H_1^{(-2m)}: x_n = -2m-1\}$, where $m = -M, -M+1, \cdots, +M$. Intuitively, this shifting process is motivated by the fact that
\begin{align}
L_{+-}^{(2m)}\left({\boldsymbol y}\right) = L_{+-}\left({\boldsymbol y}+2m{\boldsymbol h}_n\right),
\end{align}
where $L_{+-}^{(2m)}({\boldsymbol y})$ represents the LR between $H_0^{(-2m)}$ and $H_1^{(-2m)}$.

Once the LRs representing all the pairwise boundary decision probabilities in $\mathcal A$ are obtained, the posterior marginal estimation of $x_n$ can be constructed as follows: Assume a constant $p(x_n = -2M-1| {\boldsymbol y})$, we have
\begin{align}
\label{P_generate}
& p(x_n = -2m +1|{\boldsymbol y}) \nonumber \\
& = p\left(x_n = -2M-1| {\boldsymbol y}\right)\prod_{m' = M}^{m} L_{+-}^{(-2m')}\left({\boldsymbol y}\right).
\end{align}
This chain connection represented in~\eqref{P_generate} corresponds to the last layer of the neural network structure depicted in Fig. \ref{NN_structure}. 

We now consider how to efficiently train this structured neural network. As depicted in Fig.~\ref{NN_structure}, it can be observed that the essential component in this neural network is the ADNN. According to the discussion in Section~\ref{binary_integer_con}, we know the ADNN training set is based on a binary integer target. In order to create a similar training set for the ADNN with an arbitrary integer constraint, we introduce the following training set,
\begin{align*}
& \left\{\left(+1, {\boldsymbol {\tilde y}}^{(1)}_{+}\right), \left(-1, {\boldsymbol {\tilde y}}^{(1)}_{-}\right), \cdots, \left(+1, {\boldsymbol {\tilde y}}^{(K)}_{-}\right), \left(-1, {\boldsymbol {\tilde y}}^{(K)}_{-}\right) \right\}
\end{align*}
where 
\begin{align}
    {\boldsymbol {\tilde y}}^{(k)}_{+} &= {\boldsymbol y}^{(k)}-x_n^{(k)}{\boldsymbol h}_n + {\boldsymbol h}_n\\
    {\boldsymbol {\tilde y}}^{(k)}_{-} &= {\boldsymbol y}^{(k)}-x_n^{(k)}{\boldsymbol h}_n-{\boldsymbol h}_n.
\end{align}

\begin{figure}
    \centering
    \includegraphics[height = 0.4 \linewidth]{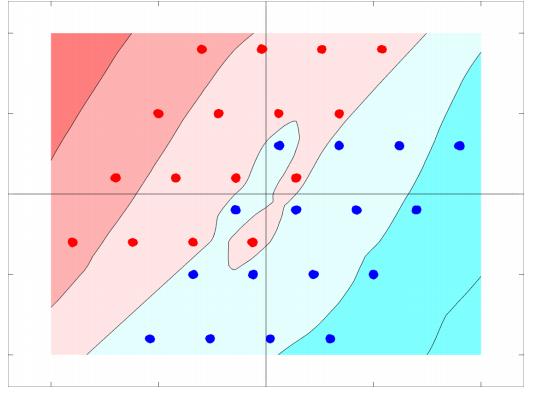}
    \caption{Decision boundary for binary integer constraint.}
    \label{Decision_boundary}
\end{figure}

A geometric illustration of this atomic decision under the general integer constraint is shown in Fig. \ref{Decision_boundary}. Suppose ${\boldsymbol y} = {\boldsymbol h}_1 {\tilde x}_1 + {\boldsymbol h}_2 x_2 + {\boldsymbol h}_3 x_3 + {\boldsymbol r}$, where ${\tilde x}_1 =\{ +1, -1\}$ and $x_2, x_3 \in \{-3, -1, +1, +3\}$. The lattice spanned by ${\boldsymbol h}_1 + {\boldsymbol h}_2 x_2 + {\boldsymbol h}_3 x_3 $ is represented by the red dots in the figure, while the blue dots represent the lattice spanned by $-{\boldsymbol h}_1 + {\boldsymbol h}_2 x_2 + {\boldsymbol h}_3 x_3 $. Due to the noise $\boldsymbol r$, the observation $\boldsymbol y$ can lie at any arbitrary location in this space. Accordingly, the ADNN generates the LR for the input $\boldsymbol y$. 

\label{binary_nn}

\subsection{Semi-Decomposed Approach}
\label{semi_learning}
Note that the aforementioned decomposed approach requires $N$ neural networks for the approximation of the marginal PDFs of the $N$ entries of $\boldsymbol x$. Furthermore, we can combine any number of entries of $\boldsymbol x$ into groups such that the marginal PDFs of more than one variable are characterized and approximated. For two entries $x_n$ and $x_{n'}$, their marginal PDF is given by
\begin{align}
    p\left(x_{n}, x_{n'} | \boldsymbol{y}\right), \qquad x_n \in {\mathcal A}, x_{n'} \in {\mathcal A}.
\end{align}
According to Bayes' rule, we then split this marginal PDF into two parts,
\begin{align}
\label{pdf2}
    p\left(x_{n}, x_{n'} | \boldsymbol{y}\right) = p\left(x_{n} | x_{n'},  \boldsymbol{y}\right)p\left(x_{n'} | \boldsymbol{y}\right).
\end{align}
Since $p\left(x_{n} | x_{n'}, \boldsymbol{y}\right) = p\left(x_{n}| \boldsymbol{y} - x_{n'}{\boldsymbol h}_{n'} \right)$, we can learn the marginal PDF of $x_n$ and $x_n'$ respectively.
Subsequently, we assemble these two marginal PDFs to generate their joint distribution by using (\ref{pdf2}). This approach also can be extended to obtain the marginal PDF of the combination of any number of variables, extending up to the joint posterior distribution. This fact reveals that the ADNN is the fundamental component in the construction of the marginal PDF of any arbitrary variable combination.

\subsection{Analytical Performance Characterization}
\label{theory}
By employing the ADNN, we can approximate the posterior PDF for any combination of variables. 
Accordingly, we can analyze how the binary decision error of the ADNN can impact the classification performance of the entire neural network. 
\begin{theorem}
Assume that the binary decision error of the ADNN is characterized by the Type-I error $\alpha$ and the Type-II error $\beta$, defined as:
\begin{align*}
\alpha = \Pr\left(L_{+-}({\boldsymbol y})<1 |x_n = +1\right)\\
\beta = \Pr\left(L_{+-}({\boldsymbol y})>1 |x_n = -1\right).
\end{align*}
Then, the decision error for the entire neural network shown in Fig. \ref{NN_structure} is bounded by
\begin{align}
\Pr(e)\leq { \bigg(\left(1-{\frac{1}{4M}}\right)(\alpha + \beta)\bigg)}^{\rho},
\end{align}
\label{thm_NN_gen}
where $\rho \in [0, 1]$, $M$ is defined in the description of the set $\mathcal A$.
\end{theorem}
Theorem~\eqref{thm_NN_gen} can be directly obtained by applying the union bound \cite{cover1999elements} to the probability of symbol error $\Pr(e)$. As an example of a numerical evaluation of this result, assume that $\alpha = 0.05$, $\beta = 0.05$, $M = 1$ and $\rho = 1$. Then, the probability of error is upper bounded by $\Pr(e)< 0.075$ which is only marginally higher than either $\alpha$ or $\beta$, the binary decision errors of the individual ADNN.
This result also indicates to what extent the performance of the algorithm would degrade with an increase in the size of the integer constraint. Asymptotically, as $M$ becomes infinitely large, the probability of the decision error for the entire neural network will tend to $({\alpha + \beta})^{\rho}$.

Having derived the relation between the probability of decision error for the atomic decision neural network (ADNN) and that for the entire neural network, we can further characterize the generalization performance of our introduced neural network by utilizing the following well-known result from PAC-Bayes learning theory. For any neural network, ~\cite{hajek2019statistical} gives the following generalization bound,
\begin{lemma}
For any training set $\mathcal{K}$ defined in~\eqref{training_set}, having a sample size $K$, a corresponding neural network ${\mathcal N}$ trained on $\mathcal{K}$ satisfies the following generalization bound, with probability at least $(1-\delta)$ for $\delta \in (0, 1)$,
\begin{align}
\label{lemma1}
\mathcal{L}\left(\mathcal N\right) \leq \mathcal{L}_{K}\left({\mathcal N}\right)+2{\mathcal R}_{K}\left({\mathcal H}\right)+\sqrt{\frac{\log \left(\frac{1}{\delta}\right)}{2 n}}.
\end{align}
\end{lemma}
We define each term in Equation~\eqref{lemma1} as follows. For a given loss function $\ell$, the \emph{risk} of $\mathcal{N}$, $\mathcal{L}(\mathcal{N})$, is defined as:
\begin{align}
    {\mathcal L}({\mathcal N}):= {\mathbb E} ( \ell({\boldsymbol x}, {\mathcal N}({\boldsymbol y})),
\end{align}
Here the expectation is taken over the joint distribution of
the input-output space. Since this joint distribution is unknown, we instead calculate the \emph{empirical} risk $\mathcal{L}_{K}(\mathcal{N})$ based on the training dataset as:
\begin{align}
    {\mathcal L}_K({\mathcal N}):= {\frac{1}{K}}\sum_k \ell({\boldsymbol x}_k, {\mathcal N}({\boldsymbol y}_k)).
\end{align}
The Rademacher complexity $\mathcal{R}_{K}(\mathcal{H})$ is a measure of the richness of the class of neural networks $\mathcal{H}$ such that $\mathcal{N} \in \mathcal{H}$, and is defined as~\cite{hajek2019statistical}: 
\begin{align}
\mathcal{R}_{K}(\mathcal{H}):=\mathbb{E}_{{\boldsymbol \epsilon}^{K},{\mathcal K}}\left[\frac{1}{K} \sup _{{\mathcal N} \in {\mathcal{H}}}\left|\sum_{k=1}^{K} \epsilon_{k} \ell\left({\boldsymbol x}_{k}, {\mathcal N}\left({\boldsymbol y}_{k}\right)\right)\right|\right],
\end{align}
where ${\boldsymbol\epsilon^{K}} := [\epsilon_1, \epsilon_2, \cdots,\epsilon_k,\cdots,\epsilon_K]$ is a vector of i.i.d. Rademacher random variables, such that each $\epsilon_k \in \{-1,+1\}$ with probabilities $\{-{\frac{1}{2}}, +{\frac{1}{2}}\}$, respectively..
\begin{lemma}
Let $\mathcal H$ represent the class of all feed-forward neural network-based classifiers with weights $\boldsymbol w$ satisfying the norm constraint $\|\boldsymbol w\| \leq B$, i.e.,: 
\begin{align*}
    \mathcal{H} := \{\left<\boldsymbol w, \cdot \right>: \|\boldsymbol w \| \leq B \}.
\end{align*}
Then, the Rademacher complexity is bounded by~\cite{hajek2019statistical}
\begin{align}
{\mathcal R}_{K}\left({\mathcal H}\right) \leq \prod_{j=1}^{J}\left(L_{j} B_{j}\right) \cdot\left({\frac{B_0R}{\sqrt{K}}} +{\frac{2R}{K}}\sqrt{J \log 2} \right),
\end{align}
where $J$ is the number of layers except the input layer, $B_j$ is the radius of the weights at the $j^{\text{th}}$ layer; the intermediate activation functions are assumed to be Lipschitz-continuous with Lipschitz constant $L_j$ and $R$ is the radius of the input training data, i.e. $R := \sqrt{\frac{1}{K}\sum_{k=1}^{K}\left\|{\boldsymbol y}_{k}\right\|^{2}}$.
\end{lemma}
Note that the loss function $\ell(\cdot)$ used in the definition of $\mathcal{L}_{K}(\mathcal{N})$ in Lemma 1 is characterized in terms of the Type-I and Type-II errors $\alpha$ and $\beta$ respectively defined in Theorem 1. Combining this relation and the known upper bound on $\mathcal{R}_{K}(\mathcal{H})$ from Lemma 2, we have the following result in Theorem 2.


\begin{theorem}
Let $\mathcal{H}_{\mathcal{A}}$ be the class of all atomic decision neural networks (ADNNs) and ${\mathcal N}_{\mathcal A}$ be a particular ADNN. With cross-entropy used as the loss function $\ell(\cdot)$, the generalization error is bounded as follows:
\begin{align*}
\Pr(e) \leq \left(\left(1-{\frac{1}{4M}}\right)\left(2{\mathcal R}_K({\mathcal H}_{\mathcal A}) + \sqrt{\frac{\log \left(\frac{1}{\delta}\right)}{2 K}} -{\mathcal L}\left({\mathcal N}_{\mathcal A}\right)\right)\right)^{\rho}
\end{align*}
with probability at least $\delta$, $\delta \in (0, 1)$.
\label{thm2}
\end{theorem}

We provide a sketch of the proof as follows: Using cross-entropy as the loss function $\ell(\cdot)$, we begin by substituting ${\mathcal L}_K({\mathcal N}_{\mathcal A}) = {\frac{1}{K}}\sum_k \mathbbm{1}(x_k = +1)\log(1-\alpha_k) +\mathbbm{1}(x_k = -1)\log(1-\beta_k)$ into (\ref{lemma1}), where $\mathbbm{1}$ represents the indicator function, $1-\alpha_k$ and $1-\beta_k$ is the ADNN outputs for $x_k=+1$ and $x_k = -1$ respectively. Recalling from Theorem 1 that the decision errors encountered during the testing stage are defined as $\alpha$ and $\beta$, we have 
\begin{align*}
{\frac{1}{K}}\sum_k \mathbbm{1}(x_k = +1)\log(1-\alpha_k) +\mathbbm{1}(x_k = -1)\log(1-\beta_k) \\
\leq {\frac{1}{2}}\log(1-\alpha) +{\frac{1}{2}}\log(1-\beta).
\end{align*}
Since $\alpha \leq 1$ and $\beta \leq 1$,  we can arrive at the inequality in Theorem~\eqref{thm2} by using the bound: $\alpha \leq -\log(1-\alpha)$ and $\beta \leq -\log(1-\beta)$.

\section{Case Study -- Classification/Symbol Detection in MIMO Systems}
\label{app}
For performance evaluation of the introduced neural network structure, we consider the task of classification/symbol detection in a multi-input-multi-output (MIMO) wireless communications system. This task of symbol detection in a wireless communications context can be viewed as a multi-label classification problem with the constraint of limited availability of training datasets, due to the physical implementation limitations of a real-world wireless communications system. In this task, $\boldsymbol H$ is the wireless transmission channel which often satisfies a certain probability distribution and $\boldsymbol y$ is the signal received at the receiver. The introduced solver is then utilized to detect the transmit symbol of interest $\boldsymbol x$, by utilizing different objective functions $f(\cdot)$.

\subsection{Comparison with Generic Neural Network Architectures}

\begin{figure}
\centering
\includegraphics[height =  0.5 \linewidth]{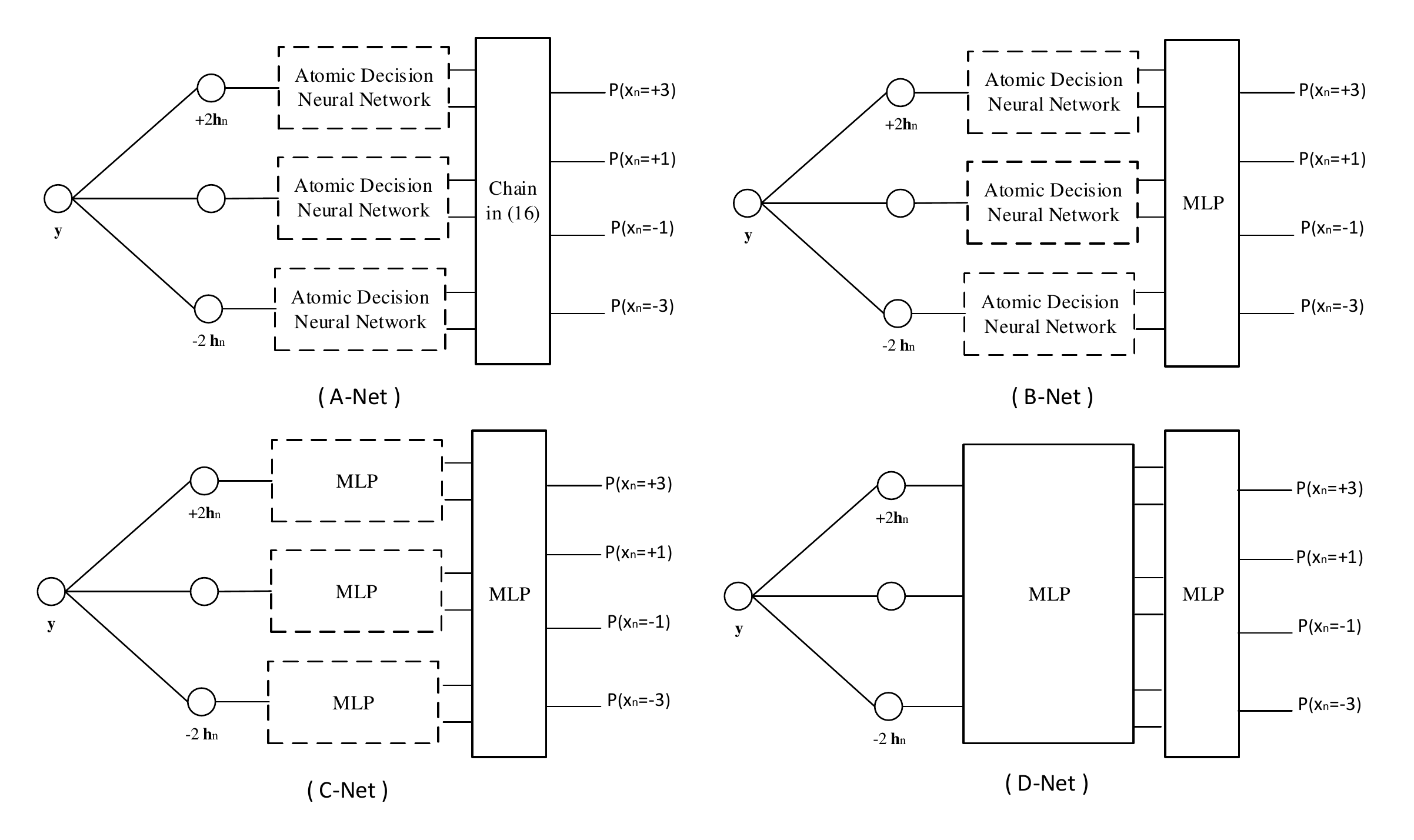}
\caption{Benchmark neural network structures for symbol detection evaluation.}
\label{Other_NN_structures}
\end{figure}

To compare the performance of our introduced neural network-based solver to other generic neural network structures, three alternate neural network structures are also developed as benchmarks. 
These four networks are respectively denoted as A-Net, B-Net, C-Net and D-Net and are shown in Fig. \ref{Other_NN_structures}. 
It can be observed that the other generic neural networks have a larger number of free parameters compared to the primary introduced method, A-Net. 
To be specific, B-Net is with a relaxation on the output layer, which is designed to gauge whether a better output combination can be learned rather than using the likelihood ratio chain as introduced in (\ref{P_generate}). 
The learning procedure for B-Net has two steps: in the first stage, we learn the atomic decision neural network as per previous discussions; in the second stage, we add a output layer to combine all these learned atomic decision neural networks and only learn the output MLP using the same training data that was used for the ADNN's learning. Note that in the second learning stage, we use multiple-class cross-entropy as the loss function. 
In C-Net, we enforce weight sharing at the same locations as the ADNNs in B-Net, and then directly learn this network. 
In D-Net, we further relax the weight sharing constraint and also learn its weights the same way as C-Net.
We first consider the case of Gaussian noise where we set $f({\boldsymbol x}) = \|{\boldsymbol x}\|_2$. 
The results comparing the generalization ability of these four neural networks are summarized in Table \ref{table_ber}, where the training set size $K$ is set as $30$. 
Subsequently, the learned neural network is tested $30000$ times by using different $\boldsymbol y$ but the same $\boldsymbol H$. 
It can be observed that the introduced structure-aware decomposition-based neural network, A-Net, has the best performance in this classification task. 
On the other hand, the structures with greater number of free parameters (in a descending order from B-Net to D-Net) have much lower generalization ability.
This clearly demonstrates the efficiency as well as the effectiveness of our introduced neural network design.

\begin{table}[!h]
\centering
\caption{Average testing and training success rate of the four neural networks using very limited training set: $K = 30$, where the objective function is $f(\cdot): = \|\cdot\|$. (one trial is counted as success when the global optimum is found)}
\begin{tabular}{ccc}
\toprule
Method & Train Success Rate & Test Success Rate \\
\hline
A-Net  &    $>0.99$   &     $0.88$ \\
B-Net  &    $>0.99$   &     $0.84$ \\
C-Net  &    $>0.99$   &     $0.70$ \\
D-Net  &    $>0.99$   &     $0.60$ \\
\bottomrule
\end{tabular}
\label{table_ber}
\end{table}

We also analyze the performance of the introduced neural network in the case where $f({\boldsymbol x}) = \sum_n \log(1 + x^2_n/\nu)$ which is parameterized by the degree of freedom parameter $\nu$. 
Accordingly, the residual noise $\boldsymbol r$ is characterized by the Student's-$t$ distribution. 
This is often the case in real-world wireless systems where $\boldsymbol r$ often represents sparse noise or sparking interference from other users whose occurrence is unknown, thus rendering the distribution of $\boldsymbol r$ to be non-Gaussian~\cite{kim2018communication}. 
The results comparing the generalization ability of the four networks using the Student's-$t$ residual model as shown in Table \ref{table_ber_t-dist}. 
This performance evaluation uses $\nu = 3$, with the training set and testing set sizes set to $K = 30$ and $30000$ respectively.

\begin{table}[!h]
\caption{Average testing and training success rate of the four neural networks using $f(\cdot) = \sum_n \log(1+(\cdot)^2_n/\nu)$ with degrees of freedom parameter $\nu = 3$.}
\centering
\begin{tabular}{ccc}
\toprule
Method & Train Success Rate & Test Success Rate \\
\hline
A-Net  &    $>0.99$  &     $0.87$ \\
B-Net  &    $>0.99$  &     $0.86$ \\
C-Net  &    $>0.99$  &     $0.70$ \\
D-Net  &    $>0.99$  &     $0.62$ \\
\bottomrule
\end{tabular}
\label{table_ber_t-dist}
\end{table}

\subsection{Theoretical Justification} 

\begin{figure}[ht]
    \centering
    \includegraphics[height = 0.5 \linewidth]{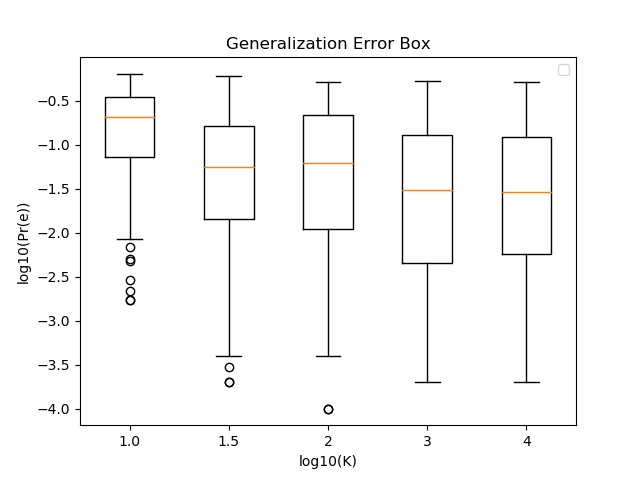}
    \caption{Generalization performance with respect to $K$ for the introduced neural network.}
    \label{boxplot}
\end{figure}

In this section, we provide experimental validation for the upper bound on the generalization error obtained in Theorem~\ref{thm2} via simulations on the introduced neural network structure A-Net. As seen in Fig.~\ref{boxplot}, the generalization error decreases with an increase in the size of the training set $K$, where the trials are conducted over 100 wireless channel realizations $\boldsymbol H$ in accordance with the $O(\frac{1}{\sqrt{K}})$ dependence evident from Theorem~\ref{thm2}.   

\subsection{Comparison with State-of-the-Art Methods}
For a complete comparison with state-of-the-art, the performance of the introduced structure-aware decomposition-based neural network (A-Net) is compared with benchmark conventional optimization methods. 
These include gradient-free optimization methods implemented from the open-source `NLOpt' package~\cite{nlopt}, such as Subplex \cite{rowan1990functional}, Nelder-Mead Simplex \cite{nelder1965simplex}, Principal Axis (PRAXIS) \cite{brent2013algorithms}, Constrained Optimization By Linear Approximations (COBYLA) \cite{powell1994direct}, Divided Rectangles algorithm (DIRECT) \cite{jones1993lipschitzian} and its locally-biased variant DIRECT-L~\cite{gablonsky2001locally}. These evaluations use a random initialization with the number of iterations set to 1000.
A-Net's performance is also compared against the genetic algorithm~\cite{genetic2007, evolutionary2015}, which was implemented using MATLAB's Optimization Toolbox. 
In addition, since the problem chosen is the specific task of symbol detection under a wireless channel, we also evaluate a state-of-the-art neural network-based MIMO symbol detection architecture, \emph{DetNet}~\cite{samuel2019learning} that is based on unfolding iteration-based optimization algorithms.
For a fair comparison, the same training and test settings are employed with \emph{DetNet} as those with A-Net.
In all of these above comparative evaluations across all trials, we set ${\mathbb E}\{ \|{\boldsymbol H}{\boldsymbol x}\|^2/ {\|{\boldsymbol H}{\boldsymbol x} - {\boldsymbol y}\|^2} \}$ as $10$ dB. We analyze two cases: where the noise $\mathbf{r}$ follows a i) Gaussian distribution and ii) a non-Gaussian distribution ($t$-distribution with $\nu=3$). The two different objective functions mentioned previously are considered as well, i.e., the L$2$-norm and the log-sum function.

The performance comparison is summarized in Table~\ref{table_diff_opt}.
We can clearly observe from Tables~\ref{table_ber}, \ref{table_ber_t-dist} and \ref{table_diff_opt} that A-Net has a higher average detection success rate when compared to either the standard optimization methods or a state-of-the-art neural network-based symbol detector such as \emph{DetNet}. This performance improvement is primarily due to the fact that we are able to incorporate the structural information in the design of A-Net.

\begin{table}[!h]
\caption{Performance comparison with non-linear optimization solvers.}
\centering
\begin{tabular}{c|c|c}
\toprule
\multirow{2}{*}{Methods} & \multicolumn{2}{l}{Symbol Detection Success Rate} \\
\cline{2-3} 
& $\|\cdot\|_2$ & $\sum_n \log(1+(\cdot)^2_n/\nu)$  \\
\hline
Sbplx& $0.87$ & $0.85$\\
\hline
Nelder-Mead & $0.89$ & $0.84$\\
\hline
PRAXIS & $0.76$ & $0.74$ \\ 
\hline
COBYLA & $0.92$ & $0.92$\\
\hline
DIRECT & $0.85$ & $0.82$ \\
\hline 
DIRECT-L & $0.83$ & $0.78$\\
\hline
Genetic Algorithm    & $0.71$             & $0.74$                      \\
\hline
\emph{DetNet}      & $0.36$                   & $0.32$               \\
\bottomrule
\end{tabular}
\label{table_diff_opt}
\end{table}

\section{Conclusion and Future Work}
\label{conclusion}
In this paper, a neural network based-approach to solve an integer-constrained programming problem was presented. The introduced neural network architecture leverages the following two features of the optimization problem (\ref{Opt}) : (i) the lattice structure of the integer constraint $\mathcal A$ (ii) the affine mapping inside the composition-type objective function $f({\boldsymbol H}{\boldsymbol x}-{\boldsymbol y})$. Moreover, the objective function is translated to its corresponding probabilistic model to generate the dataset for training the neural network. In addition, a generalization performance bound for this neural network was also derived. Experimental results verified the superior generalization ability of this method with limited training. Meanwhile, it was also observed that the introduced NN-based solver outperforms state-of-the art generic non-linear optimization solvers. Overall, the introduced structure-aware NN-based method is a promising direction towards solving NP-hard integer-constrained optimization problems.

\bibliographystyle{unsrt}
\bibliography{ref}

\end{document}